\documentclass{article}
\usepackage{spconf,graphicx}
\usepackage{multirow}
\usepackage{array}
\usepackage{arydshln}
\usepackage{amsmath}
\usepackage{amssymb}
\usepackage{graphicx}
\usepackage{subcaption}

\usepackage{enumitem}
\setlist{nosep, leftmargin=14pt}
\usepackage{mwe} 

\title{MedSegDiffNCA: Diffusion Models With Neural Cellular Automata for Skin Lesion Segmentation}
\name{Avni Mittal$^{\star}$ \qquad John Kalkhof$^{\dagger}$ \qquad Anirban Mukhopadhyay $^{\dagger}$ \qquad Arnav Bhavsar$^{\star}$}
\address{$^{\star}$ Indian Institute of Technology Mandi \\
    $^{\dagger}$ Technische Universität Darmstadt}

\begin{document}
%
\maketitle
\begin{abstract}
Denoising Diffusion Models (DDMs) are widely used for high-quality image generation and medical image segmentation but often rely on Unet-based architectures, leading to high computational overhead, especially with high-resolution images. This work proposes three NCA-based improvements for diffusion-based medical image segmentation. First, Multi-MedSegDiffNCA uses a multilevel NCA framework to refine rough noise estimates generated by lower level NCA models. Second, CBAM-MedSegDiffNCA incorporates channel and spatial attention for improved segmentation. Third, MultiCBAM-MedSegDiffNCA combines these methods with a new RGB channel loss for semantic guidance. Evaluations on Lesion segmentation show that MultiCBAM-MedSegDiffNCA matches Unet-based model performance with dice score of 87.84\% while using 60-110 times fewer parameters, offering a more efficient solution for low resource medical settings.
\end{abstract}

\section{Introduction}

Denoising Diffusion Probabilistic Models (DDPMs) \cite{ddpm, nichol2021improvedddpm} have emerged as powerful tools for image processing and generation, showing particular promise in medical image segmentation \cite{amit2021segdiff, wu2024medsegdiff, wu2024medsegdiff2}. Typically, DDPMs use the U-Net architecture \cite{zhou2018unet++} to capture multi-scale spatial dependencies, leveraging its encoder-decoder structure. However, U-Net’s large parameter count poses challenges due to the repeated use across multiple diffusion steps, leading to high computational costs.

To overcome these limitations, Neural Cellular Automata (NCA) \cite{mordvintsev2020growingnca} offers a lightweight alternative, significantly reducing parameter counts. In NCAs, each pixel is treated as a cell with an internal state that evolves based on interactions with neighboring cells, controlled by a shared neural update function. This efficiency makes NCAs particularly suitable for resource-constrained environments, with their parallelism and simplicity enabling faster training and deployment.

This work integrates NCA with DDPM for medical image segmentation and has the following salient features:
\begin{enumerate}
\item \textbf{Model Architecture: } 
We introduce three novel frameworks that target high segmentation accuracy while minimizing computational overhead.

\begin{enumerate}
    \item \textbf{Multi-MedSegDiffNCA:} A multilevel NCA architecture that refines noise estimates iteratively during backward diffusion.
    \item \textbf{CBAM-MedSegDiffNCA:} Enhances feature representation by incorporating global and spatial channel attention using the Convolutional Block Attention Module (CBAM) \cite{woo2018cbam}, marking a first for applying global attention in NCA.
    \item \textbf{MultiCBAM-MedSegDiffNCA:} Combines the benefits of multilevel NCA and CBAM, offering a more generalizable solution.

\end{enumerate}
    \item \textbf{Efficiency:} Our models achieve close to state-of-the-art performance while reducing model size by 60-110x, making them ideal for deployment in resource-limited settings.
    \item \textbf{RGB Channel Loss:} A novel loss function based on RGB channels guides the diffusion process, leveraging image semantics for better segmentation results.
\end{enumerate}

\section{Related Work}
\subsection{Denoising Diffusion Probabilistic Models}
Denoising Diffusion Models (DDMs) \cite{ddpm} have advanced image generation, showing strong performance in inpainting \cite{lugmayr2022repaint_inpainting}, segmentation \cite{amit2021segdiff, wu2024medsegdiff, wu2024medsegdiff2} etc. Ho et al. \cite{ddpm} introduced a forward diffusion process that incrementally adds noise to an image \( x_0 \) over \( T \) time steps, followed by a reverse process to reconstruct the original image:
\vspace{-1em}
\[
p_{\theta}(x_{0:T-1}|x_T) = \prod_{t=1}^{T} p_{\theta}(x_{t-1}|x_t) \quad \text{(1)}
\]
where \( p_{\theta}(x_{t-1} \mid x_t) \) represents the transition from step \( t \) to \( t-1 \), allowing recovery of clean data from noisy inputs.


SegDiff \cite{amit2021segdiff} first applied DDMs for segmentation by iteratively refining segmentation maps. MedSegDiff \cite{wu2024medsegdiff} enhanced this with a dynamic conditional encoding framework and an FF-Parser to filter out high-frequency noise. MedSegDiff-V2 \cite{wu2024medsegdiff2} further improved segmentation by integrating a transformer encoder with attention mechanisms, allowing for better focus on critical features.

\subsection{Neural Cellular Automata (NCA)}

Neural Cellular Automata (NCA) \cite{mordvintsev2020growingnca} integrate cellular automata principles with neural networks, offering a lightweight approach for image processing. NCA models treat image pixels as a grid of cells with internal states that evolve based on interactions with neighboring cells, governed by a shared update function learned by a neural network. This update function uses convolution operations to generate a perception vector, which is then processed to update cell states iteratively. Each update across the grid constitutes an NCA step, refining the image over multiple iterations while maintaining a low parameter count.

NCAs have demonstrated effectiveness in image generation \cite{otte2021generative_gannca, palm2022variationalnca}, segmentation \cite{kalkhof2023mednca, kalkhof2023m3dnca}, and inpainting \cite{tesfaldet2022attentionnca, palm2022variationalnca}, offering computational efficiency through parameter sharing. However, their local update mechanism can limit global information propagation in high-dimensional images, as it depends on the number of NCA steps. Adaptations like MED-NCA and M3D-NCA \cite{kalkhof2023mednca, kalkhof2023m3dnca} address this with multi-level architectures and patch-wise training, reducing VRAM usage. Integrating NCAs with DDPMs \cite{kalkhof2024frequency} show their potential as a Unet alternative.

\subsection{Limitiations of current work}

State-of-the-art segmentation models, typically based on UNet, have high parameter counts which result in significant computational demands and longer processing times. This complexity can be a bottleneck, especially in resource-constrained settings requiring real-time processing. NCAs offer a promising alternative by drastically reducing parameter counts, from billions to a few hundred thousand \cite{kalkhof2023mednca}. This efficiency allows NCAs to operate on smaller computational devices, reducing both, training and inference time and easing integration across various platforms, making them suitable for a wider range of applications.

\section{Methodology}

\begin{figure}[tb]
  \centering
  \includegraphics[height=4cm]{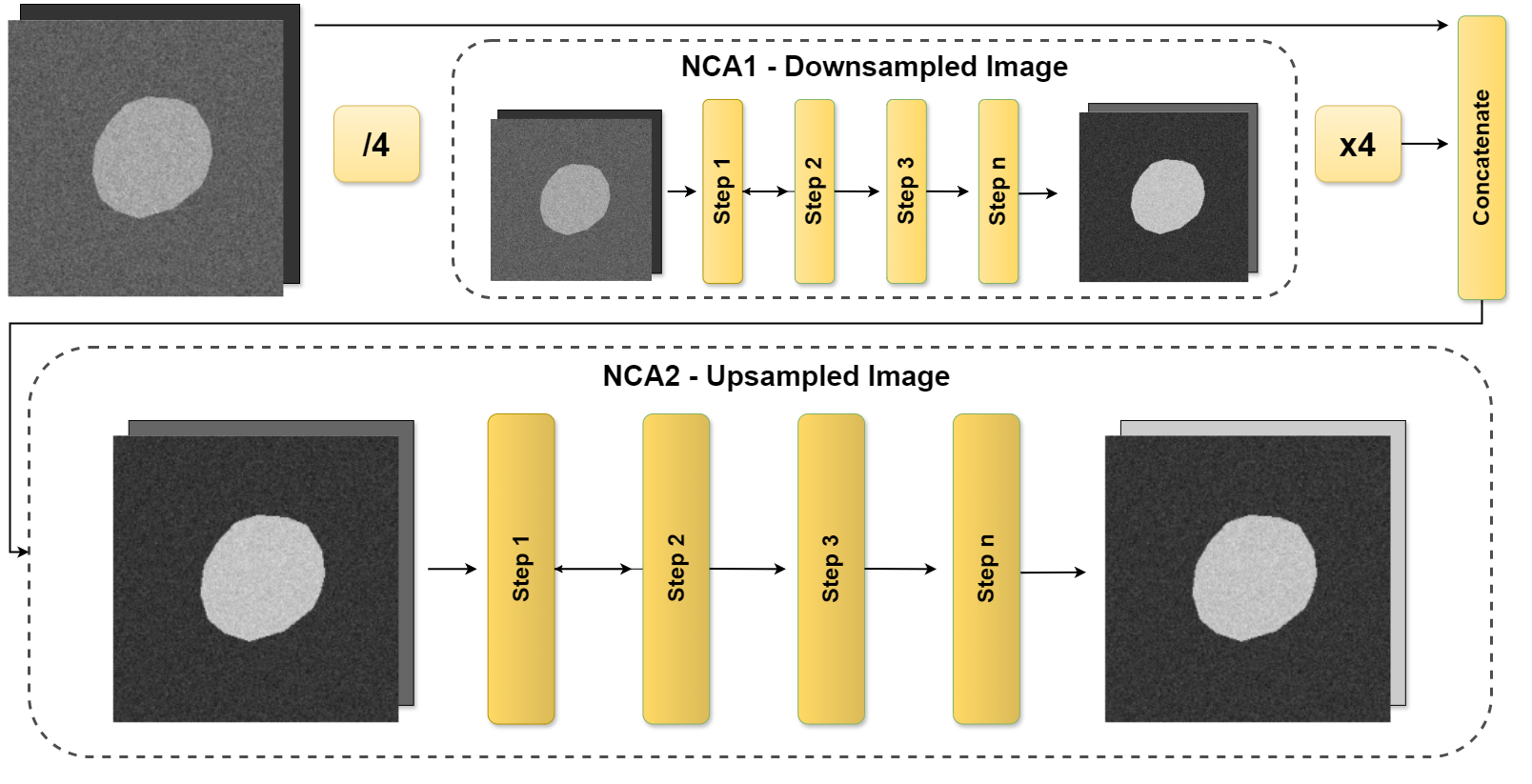}
  \caption{Multi-MedSegDiffNCA single diffusion step}
  \label{fig:multi_nca}
\end{figure}

While combining the basic NCA model \cite{mordvintsev2020growingnca} with a DDPM backend, there are challenges in efficiently propagating global context with limited NCA steps for higher-dimensional images. To address this, we explored two strategies:

\begin{enumerate}
    \item \textbf{Multilevel Pyramidal Architecture:} Facilitates global information transfer across low-dimensional representations of images.
    \item \textbf{Guiding Generation with Global Attention:} Uses attention mechanisms to ensure important global features are captured during map refinement.
\end{enumerate}

Based on these strategies, we introduce three novel NCA architectures: the Multi-MedSegDiffNCA, CBAM-MedSegDiffNCA, and MultiCBAM-MedSegDiffNCA, designed to improve segmentation quality for higher-dimensional images.

\subsection{MedSegDiffNCA} \label{medsegdiffnca}

The MedSegDiffNCA architecture replaces the Basic NCA model as the backbone for noise estimate prediction in the DDPM backend. The NCA comprises \( c \) channels, all initialized to zero. The first channel computes diffusion noise loss, while the next three channels are initialized with the RGB channels of the conditional image \( I \). The subsequent channel holds the noise-added segmentation map \( x_t \) at a specific diffusion time step, and the final channel contains the diffusion time step information \( t \):
\[
\epsilon_{\theta}(x_t, I, t) = {NCA_{\theta} [\textbf{0}, I, x_t, ... , t]} \quad \text{(2)}
\]
Here, \( \epsilon_{\theta}(x_t, I, t) \) represents the noise generation function conditioned on the raw image \( I \), diffusion time step \( t \), and current noisy segmentation map \( x_t \). The NCA model performs \( n \) iterations at each diffusion time step, with the first channel providing the predicted noise estimate for the reverse diffusion process.

\subsection{Multi-MedSegDiffNCA}

To overcome the slow local propagation of information in NCA models for larger images, we propose Multi-MedSegDiffNCA, which features multiple levels of NCAs. The first level processes downsampled images to enhance global information propagation. As representations and noise estimates are progressively upscaled, the original higher-dimensional inputs replace the downsampled inputs, allowing higher-level NCAs to leverage global embeddings. Unlike MED-NCA \cite{kalkhof2023mednca}, which refines individual image patches, our approach refines the complete noise map for the backward diffusion process.

The implementation features two levels: the original image is downsampled by a factor of 4 in the first level to produce an initial noise estimate. This estimate and the learned hidden channels are then upsampled for refinement in the second level. The refined noise estimate is used in the backward diffusion process as shown in Figure \ref{fig:multi_nca}. 
\begin{figure}[tb]
  \centering
  \includegraphics[height=4.5cm]{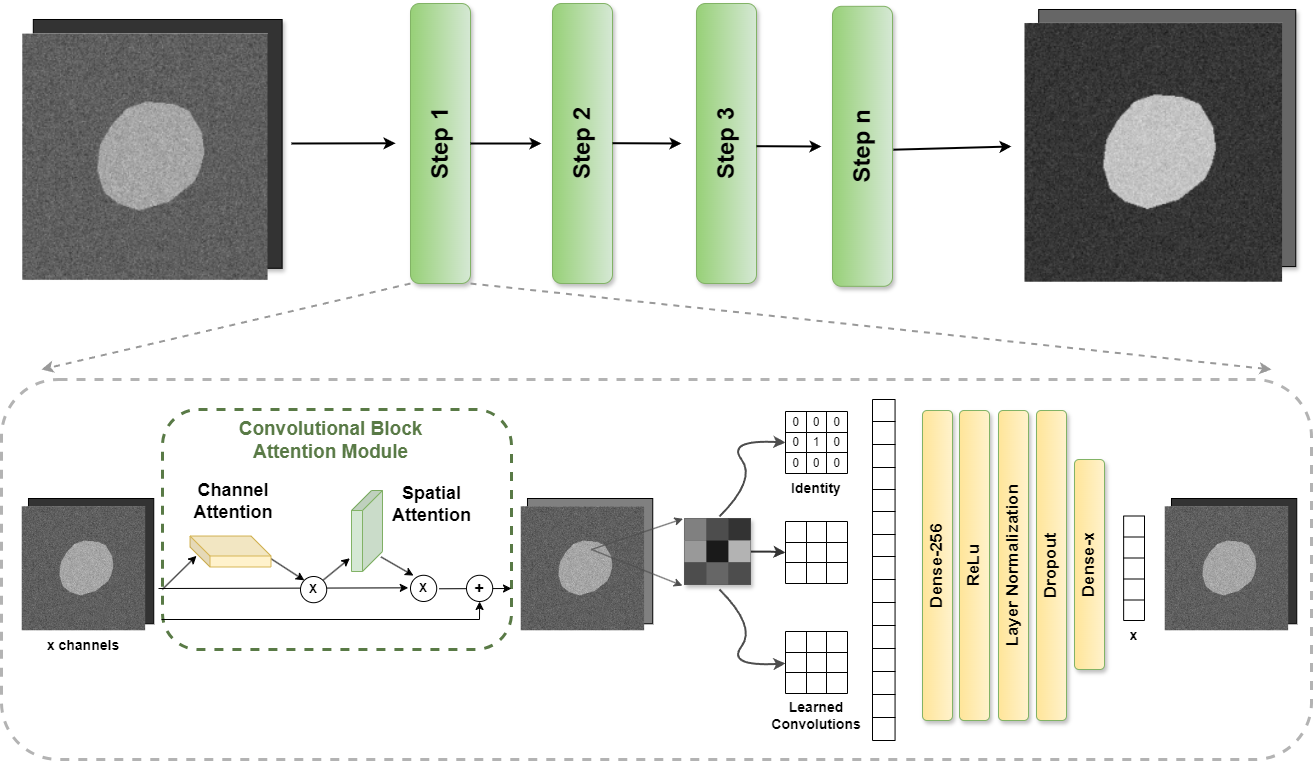}
  \caption{CBAM-MedSegDiffNCA single diffusion step}
  \label{fig:cbam_nca}
\end{figure}

\vspace{-0.17cm}
\subsection{CBAM-MedSegDiffNCA}

Basic NCA models [\ref{medsegdiffnca}] produce incomplete segmentations due to limited global image knowledge propagation. To address this, we propose integrating channel and spatial attention mechanisms into the NCA module. However, conventional attention methods like Query-Key-Value (QKV) \cite{vaswani2017attention} suffer from quadratic time complexity, causing significant computational overhead. While the ViTCA model \cite{tesfaldet2022attentionnca} combines attention with cellular automata, its local-scale operation limits global context capture.

To overcome these limitations, we incorporate the Convolutional Block Attention Module (CBAM) introduced by Woo et al. In our CBAM-MedSegDiffNCA model, the input is processed through CBAM before generating the perception vector, enabling the NCA to leverage both channel and spatial attention information while minimizing the computational costs. Figure \ref{fig:cbam_nca} illustrates the single update rule, demonstrating how the CBAM module is integrated into the NCA model.

\vspace{-0.17cm}
\subsection{MultiCBAM-MedSegDiffNCA}


\begin{figure*}[htbp]
    \centering
    \begin{subfigure}[b]{0.57\textwidth}
        \centering
        \includegraphics[width=\linewidth]{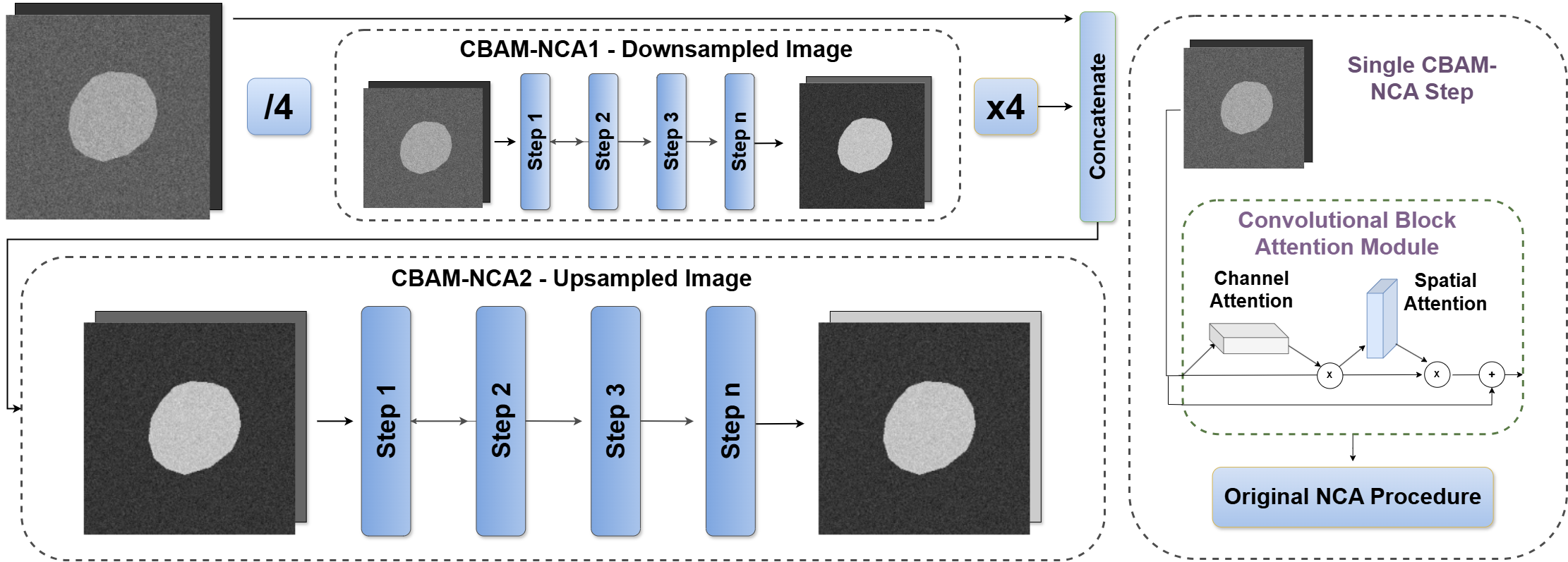} 
        \caption{MultiCBAM-MedSegDiffNCA Single diffusion step}
        \label{fig:multi_cbam_nca}
    \end{subfigure}
    \hspace{0.05\textwidth} 
    \begin{subfigure}[b]{0.35\textwidth}
        \centering
        \includegraphics[width=\linewidth]{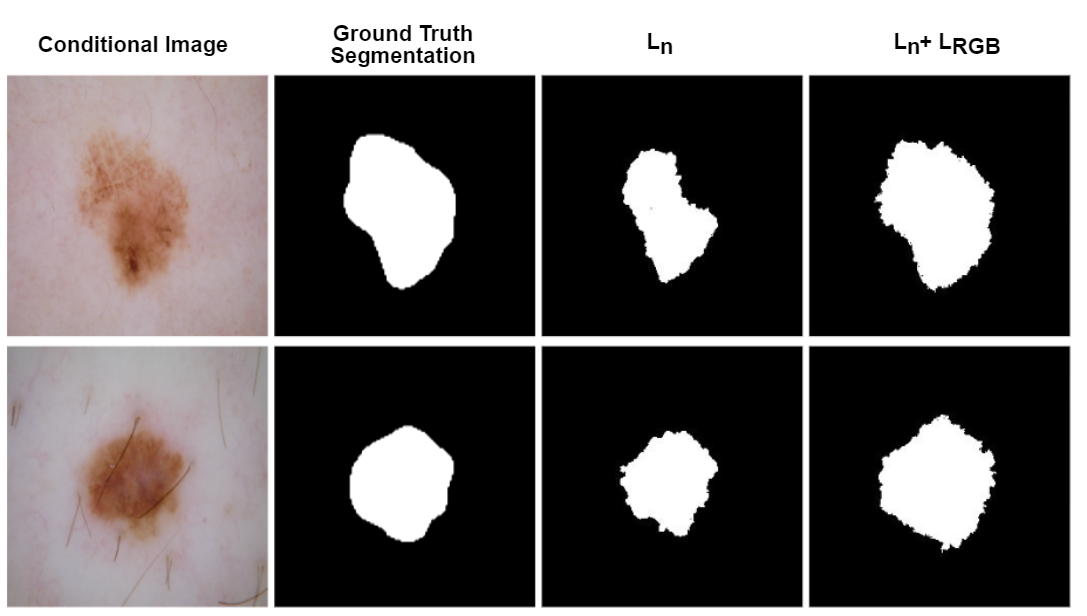} 
        \caption{Qualitative Comparison of Different Losses for Multi-MedSegDiffNCA}
        \label{fig:loss_comp}
    \end{subfigure}
    
\end{figure*}

The MultiCBAM-MedSegDiffNCA integrates the multilevel architecture of Multi-MedSegDiffNCA with the attention mechanisms of CBAM-MedSegDiffNCA, facilitating improved propagation of local information at downscaled levels while preserving global attention in segmentations. This architecture efficiently conveys global contextual understanding across multiple levels, enhancing segmentation quality by focusing on relevant features and spatial locations as illustrated in Figure \ref{fig:multi_cbam_nca}. 

\section{Experiments}

\subsection{Dataset}

For our experiments, we used the ISIC 2018 dataset \cite{codella2019skin_isic18, tschandl2018ham10000_isic18_2}, which includes 2,594 dermoscopic lesion images and corresponding segmentation masks. The masks precisely delineate lesion areas, enabling the model to learn the distinctions between cancerous and healthy skin. The default train/validation/test split is used.

\subsection{Loss}

We utilized two different loss functions in our study. Firstly, standard noise prediction loss \( L_n \) of DDPM\cite{ddpm, nichol2021improvedddpm} was used. 
\[
L_n = \mathbb{E}_{x_0, \hat{\epsilon}, t}\left[ \left\| \hat{\epsilon} - \epsilon_{\theta}\left(\sqrt{a_t}x_0 + \sqrt{1 - a_t}\epsilon, I, t\right) \right\|^2 \right]  \quad \text{(3)}
\]
To further guide the NCA model regarding the semantics of the image, we introduce an additional RGB channel loss, \( L_{RGB} \). This loss is the sum of the mean squared error between the RGB channels of the conditional raw image \( I \) modified by the NCA model and the ground truth segmentation mask \(x_0\). \(NCA[I(R)]\), \(NCA[I(G)]\) and \(NCA[I(B)]\) are the RGB channels of the raw image \( I \) modified by the NCA model respectively
\[
\begin{split}
L_{RGB} = \left\| NCA[I(R)] - (x_0)\right\|^2 + \left\| NCA[I(G)] - (x_0)\right\|^2 \\
+ \left\| NCA[I(B)] - (x_0)\right\|^2  \quad \text{(4)}
\end{split}
\]
By incorporating \( L_{RGB} \), the NCA channels are trained to extract structural information from the image, generating a preliminary segmentation mask for NCA guidance. The combined loss \( L_{total} \) used for training is:
\[
L_{total} = L_n + L_{RGB} \quad \text{(5)}
\]

\subsection{Implementation Details}
We employed a standard NCA model with a hidden size of 512 in the fully connected layer and 64 channels. Asynchronous activation was simulated by randomly activating 50\% of the cells. Input images were resized to 256x256 pixels, and each NCA model ran for 10 steps.

The model was trained end-to-end using the AdamW optimizer with an initial learning rate of \(1 \times 10^{-4}\), over 100 diffusion steps and a batch size of 8. For ensemble learning, it was run 10 times, with evaluation metrics including Dice score and Intersection over Union (IoU).

\section{Results}
Table \ref{tab:results} compares the performance of various models. While MedSegDiff and MedSegDiff-v2 achieved Dice scores of 91.3\% and IoU of 93.2\% in their original papers, these models require significantly more parameters and training time—over five times that of NCA-based models. In our experiments, we reached only 72\% Dice after training for five times longer on an NVIDIA V100 GPU, unlike \cite{wu2024medsegdiff, wu2024medsegdiff2} which used multiple A100 GPUs. Performance improved with extended training, but resource limitiations prevented full replication of the original results.

The MedSegDiffNCA model, which incorporates a basic NCA into the diffusion process, achieved a Dice score of 63.73\% and IoU of 51.19\% with just 139k parameters. Incorporating the CBAM module into MedSegDiffNCA improved the performance to 68.53\% Dice and 53.01\% IoU, highlighting the benefits of attention mechanisms. While in the above two cases, the parameter count is significantly lower, the perfrmance is also not very high. However we achieve considerably better performance with Multi-MedSegDiffNCA, with only a small increase in the number of parameters to get a Dice score of 83.9\% and an IoU of 74.5\%.

\begin{table}[htbp]
  \caption{Comparison of SOTA diffusion-based medical image segmentation and proposed NCA-based architectures}
  \centering
  \label{tab:results}
  \begin{tabular}{p{4.45cm}>{\centering\arraybackslash}p{0.6cm}>{\centering\arraybackslash}p{0.6cm}>{\centering\arraybackslash}p{1.6cm}}
  
      \hline
    \textbf{Algorithm} & \textbf{Dice} & \textbf{IoU} & \textbf{Parameters}  \\
    \hline
    MedSegDiff \cite{wu2024medsegdiff}*& 91.3 & 84.1 & 25M \\ 
    MedSegDiff-v2 \cite{wu2024medsegdiff2}*& 93.2 & 85.3 & 46M \\ 
    \hdashline
    MedSegDiffNCA & 63.73 & 51.19 & 139k   \\ 
    CBAM-MedSegDiffNCA & 68.53 & 53.01 & 206k  \\ 
    Multi-MedSegDiffNCA & 83.90 & 74.55 & 410k \\ 
    MultiCBAM-MedSegDiffNCA & \textbf{87.84} & \textbf{78.86} & \textbf{412k}   \\ 
    \hline
      \end{tabular}
    \raggedright\footnotesize{* Results from original paper.}
      
\end{table}

\begin{figure}[htbp]
  \centering
  \includegraphics[height=4.5cm]{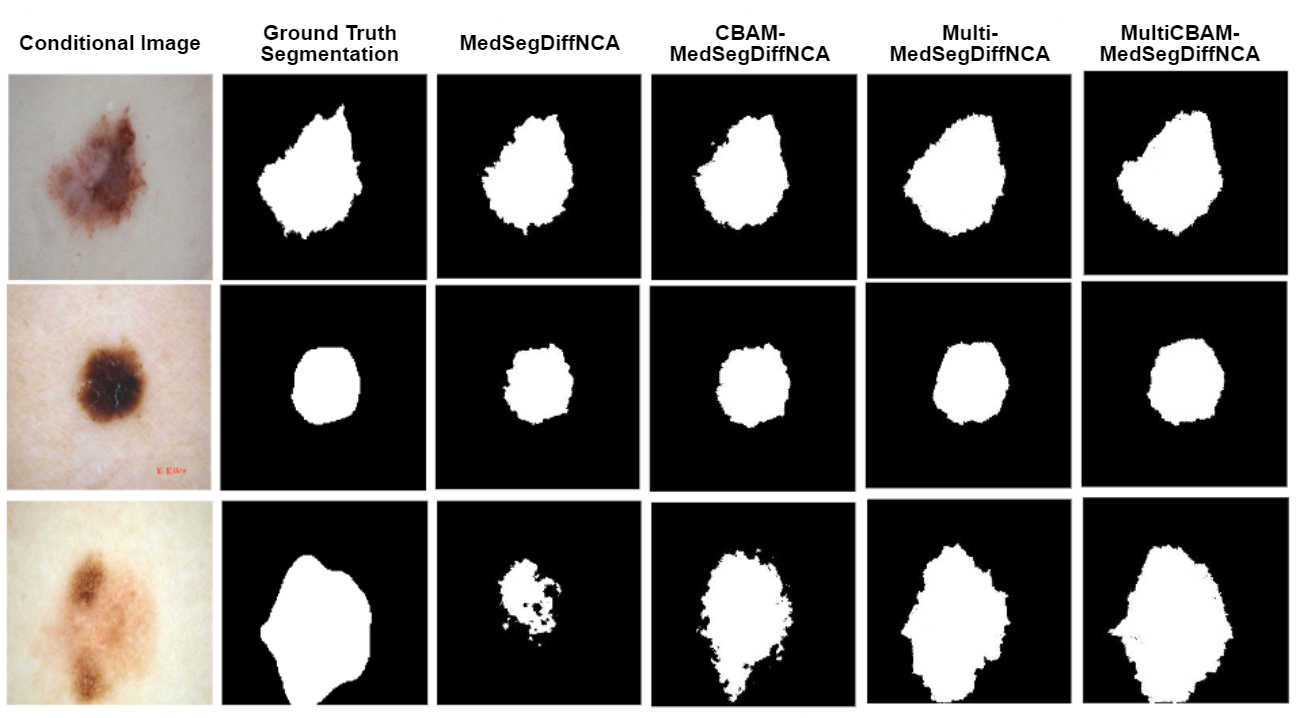}
  \caption{Qualitative Comparison of Different Methods on skin lesion segmentation}
  \label{fig:method_comp}
\end{figure}

As image resolution increases, the linear propagation of information in the NCA model becomes a bottleneck, particularly with limited NCA time steps. The multilevel architecture addresses this by enabling efficient propagation of global information in lower dimensions, leading to the improved performance of Multi-MedSegDiffNCA over CBAM-MedSegDiffNCA. The combined MultiCBAM-MedSegDiffNCA model, which integrates the multilevel NCA architecture with the CBAM module, achieves the best performance, with a Dice score of 87.54\% and an IoU of 78.86\%, while maintaining a low parameter count of 412k—over 60-110 times smaller than the MedSegDiff models.


Figure \ref{fig:method_comp} shows a qualitative comparison of segmentations from different methods. The basic NCA model \ref{medsegdiffnca} produces incomplete and patchy results, while CBAM-MedSegDiffNCA improves segmentation by incorporating channel and spatial attention mechanisms. The use of multilevel architectures further enhances segmentation quality, with MultiCBAM-MedSegDiffNCA delivering the most accurate results. Figure \ref{fig:loss_comp} compares segmentations from Multi-MedSegDiffNCA trained with standard diffusion noise prediction loss \( L_n \) versus the combined loss \( L_{total} = L_n + L_{RGB} \). Results show that \( L_n \) alone is insufficient for effective segmentation, while adding \( L_{RGB} \) significantly boosts model performance.

\section{Conclusion}

This study integrates Neural Cellular Automata (NCA) with diffusion models for medical image segmentation, introducing the MultiCBAM-MedSegDiffNCA framework. This approach combines a multilevel architecture with a Convolutional Block Attention Module (CBAM) to enhance noise prediction. We propose an RGB channel loss that conditions the model on both the original image and noisy segmentation map, improving training efficacy. Our experiments achieve state-of-the-art performance while reducing model size by over 60-110 times and decreasing training convergence time by more than 5 times. These advancements make our model particularly suitable for resource-constrained medical imaging environments.

\section{Compliance and Ethical Standards}

This research study was conducted retrospectively using human subject data made available in open access \cite{codella2019skin_isic18, tschandl2018ham10000_isic18_2}. Ethical approval was not required as confirmed by the license attached with the open access data.

\bibliographystyle{IEEEbib}
\bibliography{isbi}

\begin{thebibliography}{10}

\bibitem{ddpm}
Jonathan Ho, Ajay Jain, and Pieter Abbeel,
\newblock ``Denoising diffusion probabilistic models,''
\newblock {\em Advances in neural information processing systems}, vol. 33, pp. 6840--6851, 2020.

\bibitem{nichol2021improvedddpm}
Alexander~Quinn Nichol and Prafulla Dhariwal,
\newblock ``Improved denoising diffusion probabilistic models,''
\newblock in {\em International conference on machine learning}. PMLR, 2021, pp. 8162--8171.

\bibitem{amit2021segdiff}
Tomer Amit, Tal Shaharbany, Eliya Nachmani, and Lior Wolf,
\newblock ``Segdiff: Image segmentation with diffusion probabilistic models,''
\newblock {\em arXiv preprint arXiv:2112.00390}, 2021.

\bibitem{wu2024medsegdiff}
Junde Wu, Rao Fu, Huihui Fang, Yu~Zhang, Yehui Yang, Haoyi Xiong, Huiying Liu, and Yanwu Xu,
\newblock ``Medsegdiff: Medical image segmentation with diffusion probabilistic model,''
\newblock in {\em Medical Imaging with Deep Learning}. PMLR, 2024, pp. 1623--1639.

\bibitem{wu2024medsegdiff2}
Junde Wu, Wei Ji, Huazhu Fu, Min Xu, Yueming Jin, and Yanwu Xu,
\newblock ``Medsegdiff-v2: Diffusion-based medical image segmentation with transformer,''
\newblock in {\em Proceedings of the AAAI Conference on Artificial Intelligence}, 2024, vol.~38, pp. 6030--6038.

\bibitem{zhou2018unet++}
Zongwei Zhou, Md~Mahfuzur Rahman~Siddiquee, Nima Tajbakhsh, and Jianming Liang,
\newblock ``Unet++: A nested u-net architecture for medical image segmentation,''
\newblock in {\em Deep Learning in Medical Image Analysis and Multimodal Learning for Clinical Decision Support: 4th International Workshop, DLMIA 2018, and 8th International Workshop, ML-CDS 2018, Held in Conjunction with MICCAI 2018, Granada, Spain, September 20, 2018, Proceedings 4}. Springer, 2018, pp. 3--11.

\bibitem{mordvintsev2020growingnca}
Alexander Mordvintsev, Ettore Randazzo, Eyvind Niklasson, and Michael Levin,
\newblock ``Growing neural cellular automata,''
\newblock {\em Distill}, vol. 5, no. 2, pp. e23, 2020.

\bibitem{woo2018cbam}
Sanghyun Woo, Jongchan Park, Joon-Young Lee, and In~So Kweon,
\newblock ``Cbam: Convolutional block attention module,''
\newblock in {\em Proceedings of the European conference on computer vision (ECCV)}, 2018, pp. 3--19.

\bibitem{lugmayr2022repaint_inpainting}
Andreas Lugmayr, Martin Danelljan, Andres Romero, Fisher Yu, Radu Timofte, and Luc Van~Gool,
\newblock ``Repaint: Inpainting using denoising diffusion probabilistic models,''
\newblock in {\em Proceedings of the IEEE/CVF conference on computer vision and pattern recognition}, 2022, pp. 11461--11471.

\bibitem{otte2021generative_gannca}
Maximilian Otte, Quentin Delfosse, Johannes Czech, and Kristian Kersting,
\newblock ``Generative adversarial neural cellular automata,''
\newblock {\em arXiv preprint arXiv:2108.04328}, 2021.

\bibitem{palm2022variationalnca}
Rasmus~Berg Palm, Miguel Gonz{\'a}lez-Duque, Shyam Sudhakaran, and Sebastian Risi,
\newblock ``Variational neural cellular automata,''
\newblock {\em arXiv preprint arXiv:2201.12360}, 2022.

\bibitem{kalkhof2023mednca}
John Kalkhof, Camila Gonz{\'a}lez, and Anirban Mukhopadhyay,
\newblock ``Med-nca: Robust and lightweight segmentation with neural cellular automata,''
\newblock in {\em International Conference on Information Processing in Medical Imaging}. Springer, 2023, pp. 705--716.

\bibitem{kalkhof2023m3dnca}
John Kalkhof and Anirban Mukhopadhyay,
\newblock ``M3d-nca: Robust 3d segmentation with built-in quality control,''
\newblock in {\em International Conference on Medical Image Computing and Computer-Assisted Intervention}. Springer, 2023, pp. 169--178.

\bibitem{tesfaldet2022attentionnca}
Mattie Tesfaldet, Derek Nowrouzezahrai, and Chris Pal,
\newblock ``Attention-based neural cellular automata,''
\newblock {\em Advances in Neural Information Processing Systems}, vol. 35, pp. 8174--8186, 2022.

\bibitem{kalkhof2024frequency}
John Kalkhof, Arlene K{\"u}hn, Yannik Frisch, and Anirban Mukhopadhyay,
\newblock ``Frequency-time diffusion with neural cellular automata,''
\newblock {\em arXiv preprint arXiv:2401.06291}, 2024.

\bibitem{vaswani2017attention}
Ashish Vaswani, Noam Shazeer, Niki Parmar, Jakob Uszkoreit, Llion Jones, Aidan~N Gomez, {\L}ukasz Kaiser, and Illia Polosukhin,
\newblock ``Attention is all you need,''
\newblock {\em Advances in neural information processing systems}, vol. 30, 2017.

\bibitem{codella2019skin_isic18}
Noel Codella, Veronica Rotemberg, Philipp Tschandl, M~Emre Celebi, Stephen Dusza, David Gutman, Brian Helba, Aadi Kalloo, Konstantinos Liopyris, Michael Marchetti, et~al.,
\newblock ``Skin lesion analysis toward melanoma detection 2018: A challenge hosted by the international skin imaging collaboration (isic),''
\newblock {\em arXiv preprint arXiv:1902.03368}, 2019.

\bibitem{tschandl2018ham10000_isic18_2}
Philipp Tschandl, Cliff Rosendahl, and Harald Kittler,
\newblock ``The ham10000 dataset, a large collection of multi-source dermatoscopic images of common pigmented skin lesions,''
\newblock {\em Scientific data}, vol. 5, no. 1, pp. 1--9, 2018.

\end{thebibliography}

\end{document}